\documentclass[10pt,twocolumn,letterpaper]{article}

\usepackage{cvpr}
\usepackage{times}
\usepackage{epsfig}
\usepackage{graphicx}
\usepackage{amsmath}
\usepackage{amssymb}
\usepackage{adjustbox}
\usepackage{float}
\usepackage{placeins}
\usepackage{authblk}
\usepackage[numbers]{natbib}

\usepackage[pagebackref=true,breaklinks=true,letterpaper=true,colorlinks,bookmarks=false]{hyperref}

\cvprfinalcopy % *** Uncomment this line for the final submission

\def\cvprPaperID{2449} % *** Enter the CVPR Paper ID here

\ifcvprfinal\fi
\begin{document}

%%%%%%%%% TITLE
\title{Co-localization with Category-Consistent Features\\ and Geodesic Distance Propagation}

\author[1]{Hieu Le \thanks{hle@cs.stonybrook.edu}}
\author[3]{Chen-Ping Yu}
\author[1,2]{Gregory Zelinsky}
\author[1]{Dimitris Samaras}
%\author[2]{Author E\thanks{E.E@university.edu}}
\affil[1]{Department of Computer Science, Stony Brook University}
\affil[2]{Department of Psychology, Stony Brook University}
\affil[3]{Department of Psychology, Harvard University}
\renewcommand\Authands{ and }

\maketitle
%\thispagestyle{empty}

%%%%%%%%% ABSTRACT
\begin{abstract}
   Co-localization is the problem of localizing categorical objects using only positive set of example images, without any form of further supervision. This is a challenging task as there is no pixel-level annotations. Motivated by human visual learning, we find the common features of an object category from convolutional kernels of a pre-trained convolutional neural network (CNN).  We call these category-consistent CNN features. Then, we co-propagate their activated spatial regions using superpixel geodesic distances for localization. In our first set of experiments, we show that the proposed method achieves state-of-the-art performance on three related benchmarks: PASCAL 2007, PASCAL-2012, and the Object Discovery dataset. We also show that our method is able to detect and localize truly unseen categories, using six held-out ImagNet subset of categories with state-of-the-art accuracies. Our intuitive approach achieves this success without any region proposals or object detectors, and can be based on a CNN that was pre-trained purely on image classification tasks without further fine-tuning.    
\end{abstract}

%%%%%%%%% BODY TEXT
\section{Introduction}

\begin{figure}[t!]
\begin{center}
\includegraphics[width=0.8\linewidth]{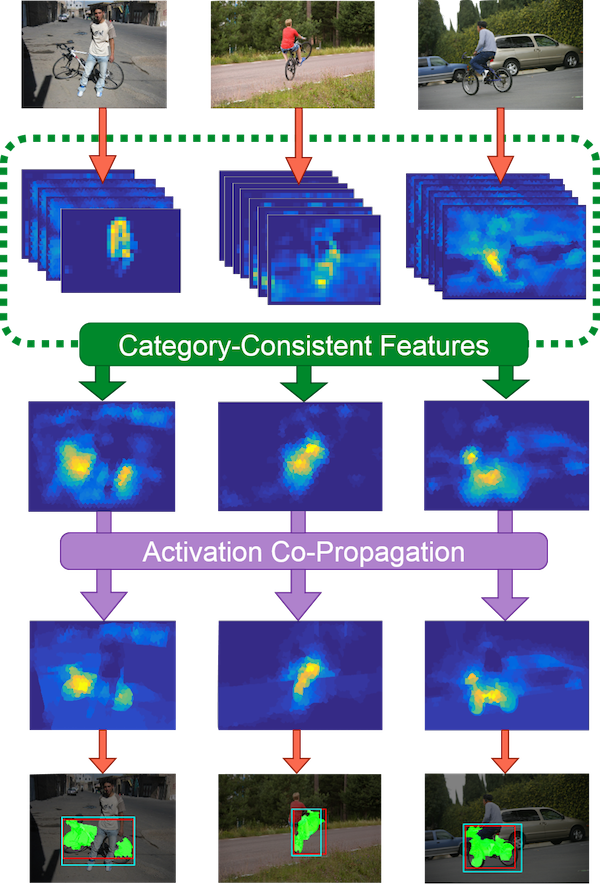}
\end{center}
   \caption{\textbf{Object co-localization with CCFs and geodesic distance co-propagation}. From a set of images containing a common object, first we find the CCFs - the group of features that consistently have high responses to the object images of the same class. The CCFs then are used to form an activation map for each image, followed by geodesic distance co-propagation to highlight the exact regions of the objects. }
\label{fig:pipeline}
\end{figure}

Convolutional neural networks (CNNs) have been widely applied in the general problem of object localization and detection. The task is to detect a target object's location and its spatial coverage in an image in the form of bounding boxes \cite{Sermanet2014,Caicedo2015,Ren2015,Redmon2016}. To localize objects from images, typically a model is given images of category exemplars for training. Critically, these training samples have precise object-level annotations, such as segmentations or bounding boxes. The models can be fine-tuned from a pre-trained network and utilize region proposals for candidate object locations \cite{Zhang2014,He2014,Zhang2015,Cinbis2016,Bency2016}, or trained end-to-end \cite{Ren2015,Tompson2015,Redmon2016}. These models have demonstrated high performance in localizing objects from learned categories, and further fine-tuning is required in order to accommodate novel object categories \cite{Oquab2015} .

Co-localization is the more challenging problem of localizing objects from only the set of positive image examples of the category without any object-level annotations. The lack of negative examples and detailed annotations hinders the use of supervised methods for the co-localization task. Recent methods typically utilize existing region proposal methods for generating a number of candidate regions for objects and object parts, followed by matching or selecting the region with the highest confidence score \cite{Joulin2014,Tang2014,Cho2015,Li2016}. However, region and object proposals are part of a research problem of its own, and have drawbacks such as lack of repeatability, reduced detection performance with a large number of proposals, and lead to difficult balance in precision and recall\cite{Hosang2015,yu2015}.

Our method, however, does not require any object proposals or object detectors to perform co-localization. %which in addition to the previously mentioned drawbacks, could also be time-consuming. 
The main idea in our work is that objects of the same class share common features or parts. Moreover, these commonalities are central to both, the category representation and the detection and localization of the object. By finding those object categorical features, their joint locations can act as a single-shot object detector. This idea is also grounded in human visual learning, where it is suggested that people detect common features from examples of the category, as part of the object-learning process \cite{Yu2016}. We do this by obtaining the CNN features of the provided set of positive images, in order to select the ones that are highly and consistently activated, which we denote as Category-Consistent CNN Features (CCFs) %following the establishment in \cite{Yu2016}. 
We then use these CCFs to discover the rough object locations, and demonstrate an effective way to co-propagate the feature activations into a stable object for precise co-localization. Figure \ref{fig:pipeline} illustrates the pipeline of our proposed framework.

%This idea is also grounded in human visual attention and learning, where Yu et al. suggested that this sort of implicit learning of an object category may occur in humans, and the detected common features would form the visual representation of the category in memory. These categorical, commonality-based features were termed the Category-Consistent Features (CCFs) \cite{Yu2016}.
%Second, the separate object features require a process to combine them into a stable object, as described in the behavioral literature as feature-cohering or binding by the focused spatial attention \cite{Rensink2000}. Based on the above, we propose to discover the rough object locations of a novel category by finding and detecting the locations of their CCFs, and then co-propagating the feature activations into a stable object for precise co-localization.       

In more detail, our approach begins with a CNN that has been pre-trained for image classification on ImageNet. Then, the images of the target category are passed through the network. We identify the last-layer convolutional filters that have highly and consistently activated feature maps as the CCFs. The CCFs' feature maps are combined into a single normalized activation probability map, where the highly activated region directly implies the rough object location, since the CCFs represent object parts or object-associated features. The CCF step allows us to bypass the need for region proposals. Then, the activation map is partitioned into superpixels and weighted by the superpixel geodesic distance into an object-likelihood map such that the responses of the object-associated features propagate over the region of the entire object. Finally, the precise object location can be obtained by placing a tight bounding box around the thresholded object-likelihood map.     

The three main contributions of this work are: 
\textbf{1.} We propose a novel CCF extraction method that can automatically highlight the rough initial object regions, which acts as a single-shot detector.
\textbf{2.} We introduce an effective method of feature co-propagation for generating a stable object region using superpixel geodesic distances on the original images. 
\textbf{3.} Our method achieves state-of-the-art performance for object co-localization on the VOC 2007 and 2012 datasets \cite{Everingham2015}, the Object Discovery dataset \cite{Rubinstein2013}, and the six held-out ImageNet subset categories. Furthermore, our framework is fully unsupervised, objects are discovered using just positive image exemplars. We are able to accurately localize objects without needing any region proposals.

%-------------------------------------------------------------------------
\section{Related work}
%Co-localization is the task of localizing the common objects within a set of images with bounding boxes. Since there is no type of annotations available, the settings of the problem is unsupervised. The lack of image-level labels and the information of the dominant object class poses a significantly harder problem than typical localization tasks. 

 Co-localization is related to work on weakly supervised object localization (WSOL) \cite{Siva11,Shi_2013_ICCV,cinbis14,Wang14,Bilen15,Ren16,Wang15} since both share the same objective: to localize objects from an image. However, since WSOL allows the use of negative examples, designing the objective function to discover the information of the object-of-interest is less challenging: WSOL-based methods achieve higher performance on the same datasets as compared to co-localization methods, due to the allowed supervised training. For instance,
\citet{Wang14} uses image labels to evaluate the discrimination of discovered categories in order to localize the objects. \citet{Ren16} adopts a discriminative multiple instance learning scheme to compensate the lack of object-level annotations to localize the objects based on the most discriminative instances. Because of the supervision that is required by those methods, it is not trivial for WSOL approaches to be directly applied to the co-localization scenarios.
%performs the object localization by firstly discovering a set of latent categories, followed by localizing the objects with the hints from the most discriminative category. The image labels are used to evaluate the discrimination of the categories. Likewise, \cite{Ren16} adopts the multiple instance learning scheme to compensate the lack of object-level annotations and utilizes MI-SVM learning to find the most discriminative instances. Such approaches cannot be directly applied on the co-localization scenario due to the unavailable of image label.

One challenge of co-localization is to define the criteria for discovering the objects without any negative examples. To fill the gap,
state-of-the-art co-localization methods such as \cite{Li2016,Tang2014,Cho2015,Joulin2014} employ object proposals as part of their object discovery and co-localization pipelines. \citet{Tang2014} use the measure of objectness~\cite{Alexe12} to generate multiple bounding boxes for each image, followed by an objective function to simultaneously optimize the image-level labels and box-level labels. Such settings allow the use of discriminative cost function~\cite{Joulin10}. This is also used in the work of co-localization on video frames~\cite{Joulin2014}. \citet{Cho2015} also starts from object proposals, their method shares the same spirit with the deformable part model~\cite{lsvm-pami} where the objects are discovered and localized by matching common object parts. Most recently, \citet{Li2016} study the confidence score distribution of a supervised object detector over the set of object proposals to define an objective function, that learns a common object detector with similar confidence score distribution. All the aforementioned methods heavily depend on the quality of object proposals.

Our work approaches the problem from a different perspective. Instead of trying to fill in the gap of the negative data and annotations that are unavailable, we find the common features shared by the objects from the positive images. Then, we use the joint locations of those features as our single-shot object detector. This allows us to bypass the need for utilizing a region or object proposal algorithm as a fist step. Our subsequent step refines the detected object features into a stable object by co-propagating their activations together. We describe the details of our 2-step approach in the following sections.

\section{Extracting Category-Consistent CNN Features}

Our proposed method consists of two main steps. The first step is to find the CCFs of a category, and obtain their combined feature map that contains aggregated CCF activations over the rough object region. Then, the CCF activations are co-propagated into a stable object using superpixel geodesic distances on the original images.

%\subsection{Extracting Category-Consistent CNN Features}
\label{sec_ccf_selection}
%An object category is defined by a set of common features that represent its different visual properties. Specifically, there exists a feature subset from a comprehensive pool of visual features to describe an object category, and we call these the category-consistent features (CCFs) of the category following \cite{Yu2016}. 

 %We aim to locate the set of representative features that can be used to estimate the object likelihood based on their feature maps.
Given a set of $n$ object images from the same class and a CNN that has been pre-trained to contain sufficient visual features, we first compute the $m$ feature maps from the $k$ last-layer convolutional kernels over the $n$ images. Then, we obtain an $m\times n$ activation matrix with each row being the activation vector $a$ of a kernel containing the maximum values of the kernel's feature maps. 

Specifically, for each kernel: $A_{i,j}  = \max(F(i,j))$, where $F(i,j)$ is the feature map of kernel $i$, given image $j$ that has been forward-passed through the CNN. The activation matrix $A$ therefore describes the max-response distributions of all kernels to all category images.

%$A_i= [a_{i,1},a_{i,2},\ldots,a_{i,n}]^T$ that contains the maximum values of the feature maps from the $n$ images. That is, $a_{i,j}  = \max(F(i,j))$, where $F(i,j)$ is the feature map of kernel $i$ given image $j$ that has been forward-passed through the CNN. Feature vector $A_i$ therefore describes the responses of kernel $i$ to that of all category instances, and we obtain a feature matrix over all kernels of the last convolutional layer. 

Our goal in this step is to identify a subset of representative kernels from the global set of candidate kernels, that contain common features from the positive images of the same class. This implies that the activation vectors of the kernels should have high values over all vector elements, since there is at least one instance of the object on every image. Conceptually, the kernels that we seek correspond to object parts, or some object-associated features. To find the CCFs, we compute the pair-wise similarities between all pairs of kernels' activation vectors, and cluster them using k-means. The kernels from the cluster with the highest mean activation correspond to the CCFs. The similarity $s_{i,j}$ between two CNN kernels $i$ and $j$, can be defined as the $L_p$ distance between two activation vectors $A_i$ and $A_j$. %is defined by Eq.\ref{equa:filterdis}
%We argue that the CCFs of the target object are the convolutional kernels that activate highly over the existence of the object. Given the

%Specifically, the $m$ kernels from the last convolutional layer of a pre-trained CNN are obtained as the set of CCF candidates, and the maximum activation of a feature map is computed as the summary activation of a kernel to an image. This forms an $m\times n$ candidate activation matrix that describes every kernel's response to that of all category instances, where a subset of them that are highly and consistently activated would be the desired CCFs. To find the CCFs from the candidate activation matrix, we compute the pair-wise similarities between all pairs of kernels and cluster them using k-means, such that the CCFs are the kernels from the cluster with the highest mean activations. The similarity $s_{k_i,k_j}$ between two CNN kernels $k_i$ and $k_j$ is defined by Eq.\ref{equa:filterdis} where $M$ is the number of images in the input set and $m_{x,y}$ is the maximum value of the activation map of kernel $x$ while taking the image $y$ as the input of the CNN.

\begin{figure}[t!]
\begin{center}
\includegraphics[width=1.0\linewidth]{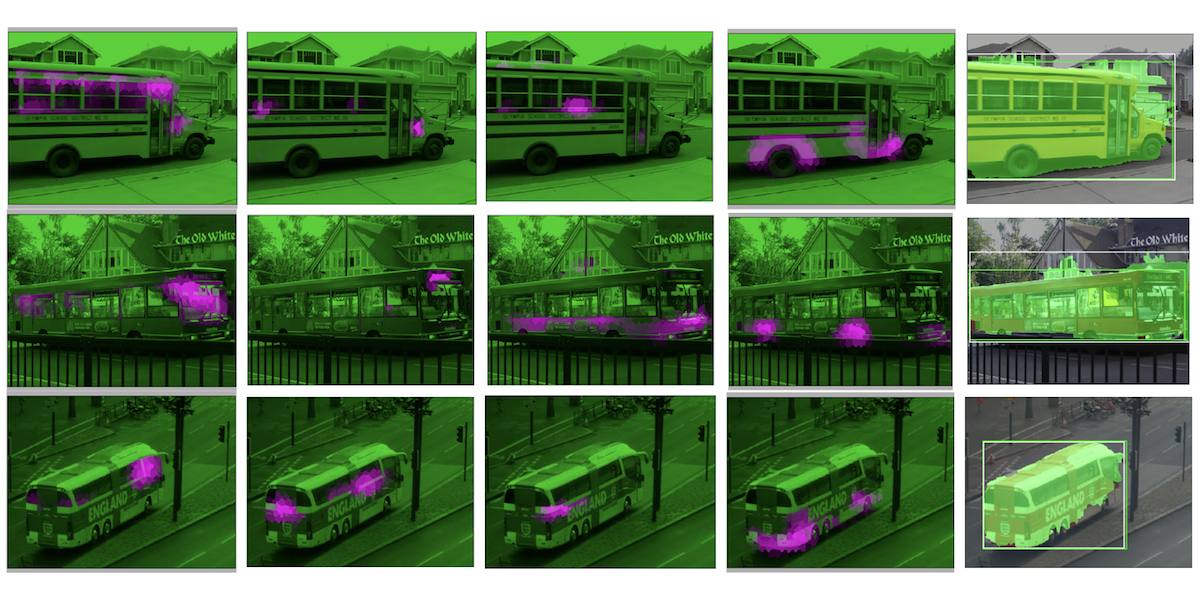}
\end{center}
   \caption{\textbf{Examples of our CCFs for \textit{bus} category.} Each row is a different example image, and each column is the activation (in violet) of a single CNN feature in the set of our CCFs. The last column shows the final co-localization results.}
\label{fig:feature}
\end{figure}

The sets of positive images are only required for the identification of the CCF kernels. The CCF kernels can then be used to generate the rough object location in an image in a single-shot: given an image from the target category, the feature maps of the CCFs are combined to form a single activation map. Since each CCF corresponds to an object part of object-associate features, the densely activated area of the activation map indicates the rough location of the target object. The final activation map is normalized into a probability map that sums to 1. In Figure \ref{fig:feature} we show the identified CCFs for the bus category, where the activated regions describe bus-related features and all fall within the spatial extent of the objects.

\section{Stable Object Completion via Co-Propagating CCF Activations}
The activation probability map from the CCFs automatically points out only the rough location of the object. It does not ensure a reliable object localization due to: 
\textbf{1.}The higher layer of a CNN does not guarantee a kernel's receptive-field size to cover the area of an entire object.
\textbf{2.}While the feature maps contain spatial information, they have unprecise object locations due to previous max-pooling layers. 
\textbf{3.}The CNN was trained discriminatively. Hence, only discriminative features of each object may be localized rather than the the whole object.

%, other than its discriminative features, w.  

%1. The output maps of the high layers of CNN barely preserve the exact locations of the objects, caused by the natural pooling and convolution scheme of CNNs. 2. The CNN features are trained to be discriminative, which does not guarantee that the object, other than its discriminative features, will be localized. 

In order to obtain the complete region that corresponds to the object, we compute geodesic distances between superpixels on the original image. In essence, the geodesic distance compactly encodes the similarity relationship between the two superpixels' image contents. The similarity is computed via object boundary detection algorithm \cite{Krähenbühl2014}. Therefore, the smaller geodesic distance between the two superpixels, the more likely that they belong to the same object. Based on this characteristic, we propose a simple  and effective method to highlight the object region from the activation probability map, that is both low-resolution and contains non-smooth feature activations.

\begin{figure*}[t!]
\begin{center}
\includegraphics[width=1.0\linewidth]{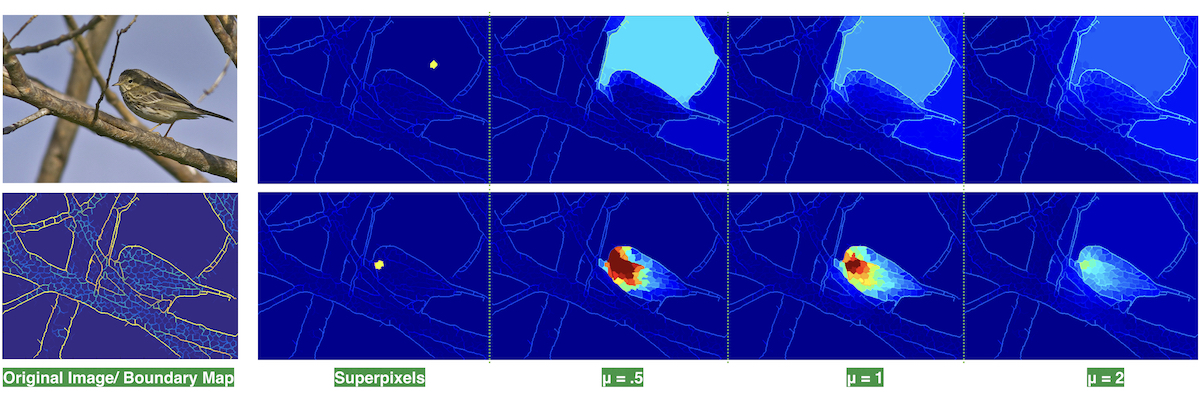}
\end{center}
\vspace{-0.2cm}
\caption{\textbf{Geodesic distance propagation.} The first column is the original image and the boundary map extracted by \cite{structured_forests}. The second column marks the position and initial activation of two selected superpixels, for illustrating the effect of our activation propagation for a single superpixel. The top superpixel is located in the background while the bottom one in located in the region of the category object (\textit{bird}). The next columns illustrate the propagated activations from the two superpixels to all other superpixels, with the controlling parameter $\mu$ set at $0.5$,$1$ and $2$, respectively.}
\label{fig:disprop}
\end{figure*}

Given an input image, we oversegment it into superpixels. The geodesic distance $d_{i,j}$ between a pair of superpixels $i,j$ is computed based on the graph built from the boundary probability map, which is done similarly as the method proposed by \citet{Krähenbühl2014}. We take the combined activation map that was obtained in the CCF identification step, and assign an energy value to each superpixel by averaging its corresponding pixel values found in the activation map. For $k$ superpixels, we denote the resulting flattened $k\times 1$ superpixel activation vector as $\mathbf{E}$. Vector $\mathbf{E}$ can be considered as the initial likelihood of each superpixel being within the object. 

Next, we perform the geodesic distance propagation to localize the object. % as the likelihood of the superpixels that belong to  similar regions, such that 
The main idea is that if two superpixels are likely to belong to the same object, then they should have similar geodesic distances and activations. This concept has been similarly adopted by various works in terms of interactive image segmentation and matting \cite{Bai2008,Grady2006}, and we find it to suit specially well for our purpose. Therefore, we seek to obtain a global activation map that has regions of highly boosted or co-propagated activations by superpixels of similar geodesic distances, mediated by some level of consistency by formulating this co-propagating mechanism into a $k \times k$ co-propagation matrix $\mathbf{W}$, such that $W_{i,j}$ is the normalized amount of co-propagation between superpixel $i$ and $j$, with a parameter $\mu$ for controlling the amount of activation diffusion:

\begin{equation}
\label{equa:W}
{W_{i,j} = \frac{{\exp(-d_{i,j}\times \mu^{-1})}}{\sum_{k=1}^N\exp(-d_{i,k}\times \mu^{-1})}}.
\end{equation}

%Next, we exploit the pair-wise similarity relationship encoded in geodesic distances to seek for a better approximation of the object's location. We first transform the geodesic distance matrix into an affinity matrix, controlled by saturation parameter $\mu$, i.e., $\exp(-d_{i,j}\times \mu^{-1})$ where $d_{i,j}$ is the geodesic distance between superpixel $i$ and $j$. We argue that, the two superpixels that are very alike, reflected in their high affinity score, should have similar activation. And thus a global, mutual activation agreement between all superpixels can be acquired if all superpixel propagate their activation scores to others, in which the sharing weights are controlled by how similar they are. We convey this propagation mechanism in Eq.\ref{equa:W}. $W$ is the $k\times k$ propagation matrix where $W_{i,j}$ is the sharing weight from superpixel $i$ to $j$. Noticeably, All values propagated from a superpixel sum to 1 to prevent the inflation. Therefore, the propagation matrix is not symmetric since each row are characterized by the distribution of the corresponding superpixel's geodesic distances to other superpixels.

Finally, we apply $W$ to the activation vector $E$ directly:

\begin{equation}
\label{equa:prop}
{E' = \mathbf{WE}},
\end{equation}

where $E'$ is the co-propagated activation vector of the image, containing the globally boosted activations of the superpixels based on their pair-wise geodesic distances to all other superpixels. This allows us to fill in each superpixel on the image with their respective values from $E'$, and normalize the co-propagated superpixel map by dividing every pixel by the max value of the map. The result is an object-likelihood map, on which we apply a global threshold to obtain the region as our final object co-localization result. Finally, a tight bounding box is placed around the maximum coverage of the thresholded regions within an image.

%The propagated energy of each superpixel is the weighted sum of the energies of all other superpixels. Importantly, the weights are characterized by the set of geodesic distances spanned from each superpixel. That is, 

Figure \ref{fig:disprop} illustrates the effects of activation propagation for two selected superpixels. The top row corresponds to the case of a background superpixel. This superpixel has small geodesic distances to a large set of other superpixels. Hence, the activation values propagated from this superpixel to others are relatively small. In the second row, we consider a superpixel corresponding to the bird. In this case, high activation values are only propagated from this superpixel to other superpixels that also correspond to the bird. Hence, we filter out the undesirable high activation of the background regions surrounding the object, while also balancing the activation of all superpixels that reside within the same object. The first column of figure \ref{fig:disprop} shows the original image and its boundary map. The second column marks the position and initial activation of the two selected superpixels. %The top superpixel is located in the background while the bottom one is located in the region of the category object (\textit{bird}).%
The next columns illustrate the activation propagating from the selected superpixel to all other superpixels with a varying degree of controlling parameter $\mu$ set at $0.5$,$1$ and $2$, where warmer values indicate higher activations. It can be seen from the third column that since the background superpixel has small geodesic distances to many other superpixels, the activations being propagated from this superpixel to others are equally small; in contrast, the activations propagated from the bottom superpixel mostly fall into its close neighboring superpixels and ones that belong to the object's region. As $\mu$ increases, the amount of propagation is more evenly and widely spread, but with a lower overall magnitude.

\section{Experiments}

We evaluate our proposed 2-step framework with different parameter settings to illustrate different characteristics of our method. We also evaluate our method on multiple benchmarks, with intermediate and final results to show the localization effects of our proposed method. In all of our experiments, we used the last convolutional layer of a VGG-19 network \cite{Simonyan14c} that was pre-trained on ImageNet \cite{russakovsky2015imagenet} as our CCF kernel pool. We used $k=5$ for kmeans clustering in the CCF identification step. The geodesic distances are computed using the Structured Forests soft boundary~\cite{structured_forests}. The control parameter $\mu$ for activation co-propagation was set at $0.5$. The final global threshold for obtaining the object region from the object-likelihood map was set at $0.25$ for all images. 

\subsection{Evaluation metric and datasets} 
We use the conventional CorLoc metric \cite{Deselaers2012} to evaluate our co-localization results. The metric measures the percentage of images that contain correctly localized results. An image is considered correctly localized if there is at least one ground truth bounding box of the object-of-interest having more than 50\% Intersection-over-Union (IoU) score with the predicted bounding box. To benchmark our method performance, we evaluate our method on three commonly used datasets for the problem of co-localization. These are VOC 2007 and 2012 \cite{Everingham2015}, and Object Discovery dataset \cite{Rubinstein2013}. For experiments on VOC datasets, we followed previous works \cite{Cho2015,Joulin2014,Li2016} that used all images on the \textit{trainval} set excluding the images that only contain the object instances annotated as \textit{difficult} or \textit{truncated}. For experiments on the Object Discovery dataset, we used the 100-image subset following \cite{Joulin13} in order to make an appropriate comparison with related methods. The ground truth bounding box for each image in the Object Discovery dataset is defined as the smallest bounding box covering all the segmentation ground truth of the object.

\begin{table*}[t!]
\centering
\begin{adjustbox}{max width=1.0\textwidth }
\begin{tabular}{c|cccccccccccccccccccc|c}
\hline
VOC  & aero          & bike          & bird          & boat          & bottle        & bus           & car  & cat           & chair        & cow           & table         & dog           & horse         & mbike         & person        & plant         & sheep         & sofa          & train         & tv            & mean          \\ \hline
\cite{Joulin2014}  & 32.8          & 17.3          & 20.9          & 18.2          & 4.5           & 26.9          & 32.7 & 41.0          & 5.8          & 29.1          & \textbf{34.5} & 31.6          & 26.1          & 40.4          & 17.9          & 11.8          & 25.0          & 27.5          & 35.6          & 12.1          & 24.6          \\
\cite{Cho2015}  & 50.3          & 42.8          & 30.0          & 18.5          & 4.0           & 62.3 & \textbf{64.5} & 42.5          & 8.6          & \textbf{49.0} & 12.2          & 44.0          & 64.1          & 57.2          & 15.3          & 9.4           & 30.9          & \textbf{34.0} & 61.6          & \textbf{31.5} & 36.6          \\
\cite{Li2016}  & \textbf{73.1} & 45.0 & 43.4          & 27.7          & 6.8           & 53.3          & 58.3 & 45.0          & 6.2          & 48.0          & 14.3          & 47.3          & \textbf{69.4} & 66.8          & \textbf{24.3} & 12.8          & \textbf{51.5} & 25.5          & 65.2          & 16.8          & 40.0          \\ \hline
ours & 56.3          & \textbf{46.75}          & \textbf{43.8} & \textbf{29.6} & \textbf{8.7} & \textbf{62.8}          & 54.6 & \textbf{74.9} & \textbf{8.8} & 42.1          & 27.7          & \textbf{50.5} & 59.2          & \textbf{71.0} & 19.3          & \textbf{13.9} & 45.4          & 31.0          & \textbf{68.9} & 9.6          & \textbf{41.2} \\ \hline
\end{tabular}
\end{adjustbox}
\vspace{0.1cm}
\caption{\textbf{CorLoc scores of our approach and state-of-the-art co-localization methods on Pascal VOC 2007 dataset.}}
\label{tab:voc07}
\end{table*}

\begin{table*}[t!]
\centering
\begin{adjustbox}{max width=1.0\textwidth }
\begin{tabular}{c|cccccccccccccccccccc|c}
\hline
VOC              & aero           & bike           & bird           & boat           & bottle         & bus            & car            & cat            & chair          & cow            & table          & dog            & horse          & mbike          & person         & plant          & sheep          & sofa           & train          & tv             & mean           \\ \hline
\cite{Cho2015} & 57.00          & 41.20          & 36.00          & 26.90          & 5.00           & \textbf{81.10} & \textbf{54.60} & 50.90          & \textbf{18.20} & 54.00          & \textbf{31.20} & 44.90          & 61.80          & 48.00          & 13.00          & 11.70          & 51.40          & \textbf{45.30} & 64.60          & \textbf{39.20} & 41.80          \\
\cite{Li2016}  & \textbf{65.70} & 57.80          & 47.90          & 28.90          & 6.00           & 74.90          & 48.40          & 48.40          & 14.60          & \textbf{54.40} & 23.90          & 50.20          & \textbf{69.90} & 68.40          & \textbf{24.00} & 14.20          & 52.70          & 30.90          & \textbf{72.40} & 21.60          & 43.80          \\ \hline
ours             & 63.06          & \textbf{67.79} & \textbf{50.39} & \textbf{41.06} & \textbf{19.73} & 79.20          & 44.50          & \textbf{74.79} & 15.27          & 40.39          & 27.42          & \textbf{68.40} & 69.36          & \textbf{74.34} & 21.17          & \textbf{14.29} & \textbf{53.37} & 39.69          & 69.41          & 15.46          & \textbf{47.45} \\ \hline
\end{tabular}
\end{adjustbox}
\vspace{0.1cm}
\caption{\textbf{CorLoc scores of our approach and state-of-the-art co-localization methods on Pascal VOC 2012 dataset.}}
\label{tab:voc12}
\end{table*}

\begin{table*}[t!]
\centering
\begin{adjustbox}{max width=1.0\textwidth }
\begin{tabular}{c|cccccccccccccccccccc|c}
\hline
VOC   & aero           & bike           & bird           & boat           & bottle         & bus            & car            & cat            & chair          & cow            & table          & dog            & horse          & mbike          & person         & plant          & sheep          & sofa           & train          & tv             & mean           \\ \hline
\cite{Siva11} & 42.40          & 46.50          & 18.20          & 8.80           & 2.90           & 40.90          & 73.20          & 44.80          & 5.40           & 30.50          & 19.00          & 34.00          & 48.80          & 65.30          & 8.20           & 9.40           & 16.70          & 32.30          & 54.80          & 5.50           & 30.38          \\
\cite{Shi_2013_ICCV} & 67.30          & 54.40          & 34.30          & 17.80          & 1.30           & 46.60          & 60.70          & 68.90          & 2.50           & 32.40          & 16.20          & 58.90          & 51.50          & 64.60          & 18.20          & 3.10           & 20.90          & 34.70          & 63.40          & 5.90           & 36.18          \\
\cite{cinbis14}  & 56.60          & 58.30          & 28.40          & 20.70          & 6.80           & 54.90          & 69.10          & 20.80          & 9.20           & 50.50          & 10.20          & 29.00          & 58.00          & 64.90          & 36.70          & 18.70          & 56.50          & 13.20          & 54.90          & 59.40          & 38.84          \\
\cite{Wang15} & 37.70          & 58.80          & 39.00          & 4.70           & 4.00           & 48.40          & 70.00          & 63.70          & 9.00           & 54.20          & \textbf{33.30} & 37.40          & \textbf{61.60}          & 57.60          & 30.10          & 31.70          & 32.40          & \textbf{52.80} & 49.00          & 27.80          & 40.16          \\
\cite{Bilen15}  & 66.40          & 59.30          & 42.70          & 20.40          & 21.30          & \textbf{63.40} & \textbf{74.30} & 59.60          & 21.10          & 58.20          & 14.00          & 38.50          & 49.50          & 60.00          & 19.80          & 39.20          & 41.70          & 30.10          & 50.20          & 44.10          & 43.69          \\
\cite{Ren16} & 79.20          & 56.90          & 46.00          & 12.20          & 15.70          & 58.40          & 71.40          & 48.60          & 7.20           & \textbf{69.90} & 16.70          & 47.40          & 44.20          & \textbf{75.50} & \textbf{41.20} & \textbf{39.60} & 47.40          & 32.20          & 49.80          & 18.60          & 43.91          \\
\cite{Wang14} & \textbf{80.10} & \textbf{63.90} & \textbf{51.50} & 14.90          & \textbf{21.00} & 55.70          & 74.20          & 43.50          & \textbf{26.20} & 53.40          & 16.30          & 56.70          & 58.30          & 69.50          & 14.10          & 38.30          & \textbf{58.80} & 47.20          & 49.10          & \textbf{60.90} & \textbf{47.68} \\ \hline
ours & 56.3          & 46.75          & 43.8 & \textbf{29.6} & 8.7 & 62.8          & 54.6 & \textbf{74.9} & 8.8 & 42.1          & 27.7          & 50.5 & 59.2          & 71.0 & 19.3          & 13.9 & 45.4          & 31.0          & \textbf{68.9} & 9.6          & 41.2 \\ \hline
\end{tabular}
\end{adjustbox}
\vspace{0.1cm}
\caption{\textbf{CorLoc scores of our approach and state-of-the-art weekly-supervised-object-localization methods on Pascal VOC 2007 dataset.}}
\label{tab:week}
\end{table*}

\begin{table}[t!]
\centering
\begin{tabular}{c|ccc|c}
Methods          & Airplane       & Car            & Horse          & Mean           \\ \hline
\cite{Kim2011}            & 21.95          & 0.00           & 16.13          & 12.69          \\
\cite{Joulin10}            & 32.93          & 66.29          & 54.84          & 51.35          \\
\cite{Joulin12}            & 57.32          & 64.04          & 52.69          & 58.02          \\
\cite{Joulin13}            & 74.39          & 87.64          & 63.44          & 75.16          \\
\cite{Joulin2014}            & 71.95          & 93.26          & 64.52          & 76.58          \\
\cite{Cho2015} & 82.93          & \textbf{94.38} & 75.27          & 84.19          \\ \hline
Ours             & \textbf{84.15} & \textbf{94.38} & \textbf{78.49} & \textbf{85.67}\\\hline
\end{tabular}
\vspace{0.2cm}
\caption{\textbf{Experiment on Object Discovery Dataset.} Highest performances are labeled in bold.}
\label{tab:OD}
\end{table}

\subsection{Comparison to state-of-the-art co-localization methods}

We first evaluate our method on the 100-image subset of Object Discovery dataset which contains objects of three classes, namely \textit{airplane}, \textit{car}, and \textit{Horse}. There are 18, 11, and 7 noisy images in each class, respectively. Table \ref{tab:OD} reports the co-localization performance of our approach in comparison with the state-of-the-art methods on image co-localization \cite{Joulin10,Joulin13,Joulin12,Joulin2014,Cho2015,Li2016}. In this small scale setting, our method outperformed other methods in both in individual object classes and overall.

\begin{figure}[t!]
\begin{center}
\includegraphics[width=1.0\linewidth]{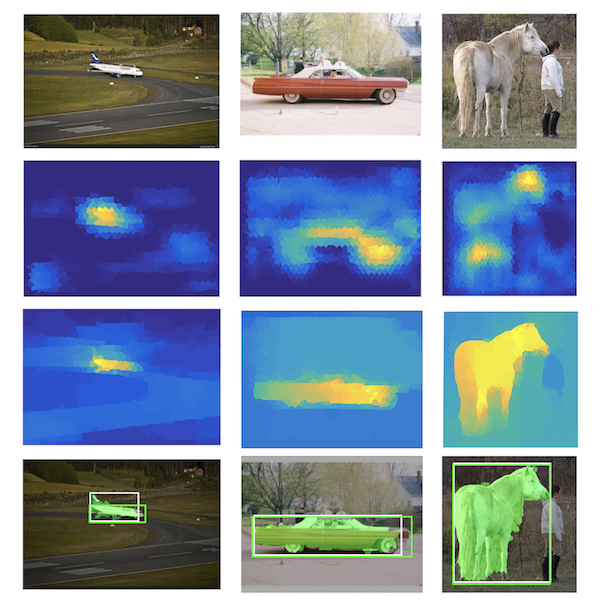}
\end{center}
   \caption{\textbf{Object co-localization results on Object Discovery dataset.} From the top row to the bottom row: input image, combined activation map from the identified CCFs, propagated object-likelihood map, and resulting bounding boxes. We also depict in green the pixels with the object-region that's predicted by our method. Our predicted bounding boxes are colored as white while the ground truth bounding boxes are green.}
\label{fig:OD}
\end{figure}

Three examples of our co-localization approach on the Object Discovery dataset are illustrated in figure \ref{fig:OD}. The second row shows the combined activation map from the set of identified CCFs, that acted as our single-shot object detector. It is apparent that the combined activation maps already provided object estimates that were quite accurate to the location of the actual object in the images, with different parts of each object getting high values such as the tail of the airplane, the wheel of the car, or head and tail of the horse. All these values were co-propagated based on the superpixel geodesic distances, resulted in the images shown in the third row. The co-propagated object-likelihood maps on the third row show that the sporadic activations on the background have been smoothed out evenly, and that the non-smooth object parts and their associated activations have been boosted and completed into complete and stable objects with significantly higher activation magnitudes. This shows that our two-step framework was able to generate informative single-shot object detection using the CCFs, and the subsequent stable object region via activation co-propagation.

%The high activation background regions surrounding the objects in the second row activation maps are filtered out since they have to co-share their values to the rest of the background while the object regions preserve high values since the whole objects region receives the high shared values from multiple parts of the objects.

%%%%%%%%%%%%%%%%%%%%%%%

The PASCAL VOC 2007 and 2012 datasetes both contains realistic images of 20 object classes with significantly larger numbers of images per class. These datasets are more challenging than the Object Discovery dataset due to the diversity of viewpoints, and the complexity of the objects. Table \ref{tab:voc07} reports our performance on the VOC 2007 dataset. Our approach outperforms the state-of-the-art method \cite{Li2016} by 1.2\% on average, and acquires the highest scores for 11 out of 20 classes. Results on VOC 2012 dataset, that has twice the number of images than the VOC 2007 dataset, are shown in table \ref{tab:voc12}. Our method also achieves significantly better results than state-of-the-art methods with an 3.65\% increase on average, acquiring the highest scores for 10 out of 20 classes.

We also compare our method with state-of-the-art weekly-supervised object localization methods \cite{Siva11,Shi_2013_ICCV,cinbis14,Wang14,Bilen15,Ren16,Wang15}, which is summarized in table \ref{tab:week}. While our results did not achieve the overall state-of-the-art, our approach was able to outperform the methods in 3 of the 20 classes, without using any negative examples or any form of supervised learning.

\subsection{Category-consistent CNN features selection analysis}
In this section, we provide an analysis to justify our CCF selection method, where we conducted additional experiments with the same configurations but using different subsets of CNN features for the initial object detection step. After clustering the last-layer CNN kernels based on their image-level activations, these clusters were sorted in a decreasing order by the clusters' average activations. We then obtained the rough object locations using individual clusters and thresholded on those maps directly to obtain the object locations. Their respective average CorLoc scores on the VOC 2007 and 2012 dataset are reported in table \ref{tab:cluster2}. 

\begin{table}[b!]
\centering
\begin{tabular}{c|ccccc}
\hline
Dataset & 1st   & 2nd   & 3rd   & 4th   & 5th   \\ \hline
VOC07   & 41.24 & 38.85 & 31.73 & 29.62 & 23.12 \\
VOC12   & 47.45 & 42.35 & 37.70 & 33.91 & 25.76 \\ \hline
\end{tabular}
\vspace{0.2cm}
\caption{\textbf{Co-localization performance of our method on the VOC 2007 and 2012 dataset.} Each column indicates which top cluster was taken as the CCF cluster (out of 5 total clusters), and the corresponding average CorLoc score (\%) by using the selected cluster of features for co-localization.}
\label{tab:cluster2}
\end{table}

\begin{table*}[t!]
\centering
\begin{adjustbox}{max width=1.0\textwidth }
\begin{tabular}{c|cccccccccccccccccccc}
\hline
Cluster & aero & bike & bird & boat & bottle & bus  & car  & cat  & chair & cow  & table & dog  & horse & mbike & person & plant & sheep & sofa & train & tv   \\ \hline
1st     & 6.3  & 4.5  & 4.1  & 3.7  & 8.0    & 7.6  & 4.1  & 3.7  & 22.9  & 6.8  & 8.4   & 5.7  & 4.1   & 5.3   & 16.2   & 9.2   & 5.7   & 20.1 & 10.4  & 7.6  \\
2nd     & 18.6 & 16.2 & 8.8  & 19.7 & 29.5   & 17.4 & 20.5 & 17.4 & 38.3  & 12.7 & 36.5  & 6.1  & 17.8  & 23.6  & 13.5   & 24.0  & 5.7   & 11.5 & 22.1  & 19.7 \\
3rd     & 30.7 & 9.6  & 13.5 & 28.3 & 10.0   & 27.5 & 32.4 & 23.6 & 7.2   & 24.4 & 8.0   & 22.9 & 13.5  & 30.7  & 26.0   & 24.0  & 20.9  & 32.4 & 26.2  & 30.3 \\
4th     & 29.5 & 43.0 & 40.6 & 24.6 & 37.5   & 28.7 & 28.7 & 23.6 & 20.1  & 34.4 & 32.8  & 38.1 & 35.2  & 29.7  & 30.5   & 28.5  & 38.3  & 26.0 & 25.2  & 28.5 \\
5th     & 15.0 & 26.8 & 33.0 & 23.6 & 15.0   & 18.8 & 14.3 & 31.6 & 11.5  & 21.7 & 14.3  & 27.3 & 29.5  & 10.7  & 13.9   & 14.3  & 29.5  & 10.0 & 16.2  & 13.9\\ \hline
\end{tabular}
\end{adjustbox}
\vspace{0.1cm}
\caption{\textbf{The number of kernels (\%) per cluster. }The percentage of kernels in each cluster based on the activation vectors described in Section \ref{sec_ccf_selection}. The clusters were formed by using K-mean with $k=5$, and the kernels were taken from the last convolutional layer of VGG-19, on each class of the VOC 2012 dataset.}
\label{tab:nclusters}
\end{table*}

\begin{figure}[t!]
\begin{center}
\includegraphics[width=1.0\linewidth]{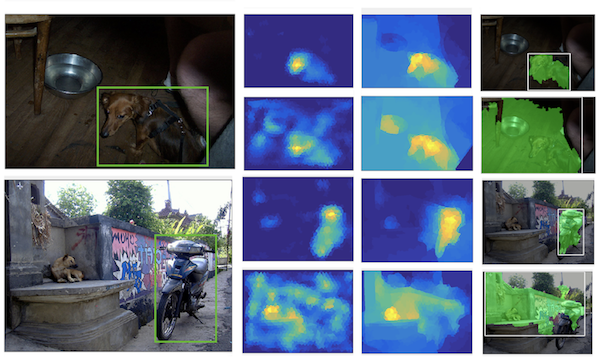}
\end{center}
   \caption{\textbf{Two examples illustrating the effect of our feature selection method.} For each image, the first row is the results of our method when using the first cluster and the second row is the results of our method when using the third cluster. The green bounding box is the ground truth and the predicted bounding boxes are colored as white, the predicted object regions are masked as green.}
\label{fig:cluster}
\end{figure}

The table shows that the co-localization performance followed in the exact order of the clusters based on their average activations, suggesting that the most representative features were indeed members of the top cluster. The visualized examples are shown in figure \ref{fig:cluster}, with an image from the \textit{dog} and \textit{motorbike} category, respectively. For each image, the first row is the results of our method when using the first cluster (ranked by their average activation) and the second row shows the results of our method when using the third cluster. It is clear that the combined activation maps from the third cluster failed to detect and estimate the object locations, and ultimately lead to incorrect object localization results. This indicates that the selection of the top cluster is essential, and the CCFs could not be chosen arbitrarily.

We furthermore validate our feature selection method by evaluating the performance of our approach when the CCFs were identified from more than one cluster, and the results are shown in table \ref{tab:cluster1}. This experiment shows that large amount of kernels do not provide enough object specificity, and therefore resulting in a similar performance decline as in table \ref{tab:cluster2}, such that performance decreases when more clusters were added.

%The results are very similar to table \ref{tab:cluster2}, but instead of using a single cluster, we also include all the features from the previous ones, i.e., the last column reports the evaluation score when all the features are selected to localize the object. As can be seen from the table, the co-localization performance also decrease gradually as more features are added. %The changes however are less pronounce than that of the case of Table.\ref{tab:cluster2} since there are always \textit{good} features in the set of considering features.

\begin{table}[h!]
\centering
\begin{adjustbox}{max width=1.0\textwidth }
\begin{tabular}{c|ccccc}
\hline
Dataset & 1   & 1-2   & 1-3   & 1-4   & 1-5   \\ \hline
VOC07   & 41.24 & 41.00 & 38.22 & 34.80 & 33.25 \\
VOC12   & 47.45 & 46.68 & 44.14 & 40.65 & 39.33 \\ \hline
\end{tabular}
\end{adjustbox}
\vspace{0.2cm}
\caption{\textbf{Co-localization performance of our methods on the VOC 2007 and 2012 dataset.} Each column indicates how many top clusters were taken as the CCF clusters (out of 5 total clusters), and the corresponding average corLoc score (\%) by using the selected clusters of features for co-localization. For example, the last column indicates that all available features were used for co-localization.}
\label{tab:cluster1}
\end{table}

We also report the number of kernels per cluster in table \ref{tab:nclusters}. Looking at table \ref{tab:nclusters} and table \ref{tab:voc12} together, they suggest that there are multiple classes that require just a small amount of kernels in order to be localized decently. For example, the \textit{cat} class used less than 20 kernels (3.7\% of the 512 total kernels) to achieve the state-of-the-art CorLoc score of 74.9\% in VOC 2012 dataset. In general, the table shows that the first cluster only contains a small number of kernels, and the results suggest that these small number of specific CCFs are sufficient for the co-localization performed on the three benchmark datasets.

%We evaluate the effectiveness of our CCFs by evaluate our method on VOC 2012 dataset with different configurations. We choose \textit{K=10} and report the CorLoc scores when choosing each individual cluster to generate the initial object-likelihood map.
\subsection{Geodesic distance co-propagation analysis}

Geodesic distance acts as a refinement step in our pipeline to get rid of the background as well as boosting the activation within the object region.
To evaluate the effect of geodesic distance co-propagation, we simply compared the performances of our method on VOC 2007 and 2012 datasets with and without geodesic distance co-propagation, and the results are reported in table \ref{tab:Geo}. The results show that geodesic distance co-propagation significantly improved the co-localization accuracy by more than 6\% in absolute CorLoc score for both dataset, which means it is an important process subsequent to our initial CCF object detection step.

\begin{table}[h]
\centering
\begin{tabular}{c|cc}
\hline
Dataset & Without GDP & With GDP \\ \hline
VOC07   & 34.45    & 41.24     \\
VOC12   & 40.97    & 47.45       \\ \hline
\end{tabular}
\vspace{0.2cm}
\caption{\textbf{Co-localization performance of our methods on VOC 2007 and 2012 dataset with and without using geodesic distance propagation.} Using geodesic distance co-propagation increased the average CorLoc score (\%) of our approach by 6.79\% in CorLoc for VOC 2007 dataset, and 6.48\% in CorLoc for VOC 2012 dataset.}
\label{tab:Geo}
\end{table}

\subsection{Unseen classes from ImageNet subset}
Our VGG-19 model was pre-trained on ImageNet's 1000 classes. While VOC 2007 and 2012 are different datasets from ImageNet, there are significant overlaps between the object categories in VOC and ImageNet. For example, the "motorbike" class of VOC datasets is equivalent to the "moped" class of ILSVRC 2012 dataset. 

\begin{table*}[t!]
\centering
\begin{tabular}{c|cccccc}
\hline
ImageNet & chipmunk       & rhino          & stoat          & racoon         &rake           & wheelchair \\ \hline
\# of images (Full)  & 307      & 213   & 237   & 1301   & 466  & 420        \\
\# of images (Smaller from \cite{Li2016})  & 159      & 89   & 159   & 103   & 146  & 174        \\ \hline
\end{tabular}
\vspace{0.2cm}
\caption{Number of images in each of the 6 subsets selected from ImageNet dataset, collected by us directly from the ImageNet website (Full),  and the smaller set by \citeauthor{Li2016}\cite{Li2016} (Smaller from \cite{Li2016}).}
\label{nim}
\end{table*}

\begin{table*}[t!]
\centering
\begin{tabular}{c|cccccc|c}
\hline
ImageNet & chipmunk       & rhino          & stoat          & racoon         & rake           & wheelchair     & mean           \\ \hline
\cite{Cho2015}      & 26.60          & 81.80          & 44.20          & 30.10          & 8.30           & 35.30          & 37.72          \\
\cite{Li2016}       & 44.00          & 81.80          & 67.30          & 41.80          & \textbf{14.50} & 39.30          & 48.12          \\ \hline
Ours     & \textbf{72.33} & \textbf{88.76} & \textbf{75.59} & \textbf{83.16} & 11.64          & \textbf{49.53} & \textbf{63.50} \\ \hline
\end{tabular}
\vspace{0.2cm}
\caption{CorLoc scores \cite{Deselaers2012} (\%) of our approach and state-of-the-art co-localization methods on the 6 smaller subsets of ImageNet collected by \citeauthor{Li2016} \cite{Li2016}.}
\label{selectedset}
\end{table*}

\begin{table*}[t!]
\centering
\begin{tabular}{c|cccccc|c}
\hline
ImageNet & chipmunk & rhino & stoat & racoon & rake  & wheelchair & mean  \\ \hline
Ours     & 76.87    & 87.79 & 80.59 & 74.10  & 50.64 & 54.05      & 70.67 \\ \hline
\end{tabular}
\vspace{0.2cm}
\caption{CorLoc scores \cite{Deselaers2012} (\%) of our approach and state-of-the-art co-localization methods on the 6 full subsets of ImageNet.}
\label{fullset}
\end{table*}
Table \ref{selectedset} reports the co-localization performance of our method on the 6 smaller subsets in comparison with the two state-of-the-art methods. The table shows that our method outperformed \cite{Li2016} and \cite{Cho2015} by a large margin on average, and all but one individual classes. %We have a slightly worst CorLoc score on "rake" than that of method \cite{Li2016}. However, all three methods seem to not perform really well on this set of images, varied from only 8.3\% to 14.5\%, possibly due to the lack of data that makes the algorithms fail to discover the common patterns of the object. %We visualize some successful co-localization results in Figure .

The six subsets of the ImageNet dataset, chosen by \citeauthor{Li2016} \cite{Li2016}, are held-out categories from the 1000-label classification task, which means they do not overlap with the 1000 classes used to train VGG-19. We show that our method is generalizable to truly novel object categories with the six held-out ImageNet subset classes. The images and the corresponding bounding box annotations were downloaded from ImageNet website \cite{ImageNet}, and
Table \ref{nim} shows the numbers of images in each class with available bounding box annotations when we accessed the ImageNet dataset website. It is noticeable that paper \cite{Li2016} used an even smaller set with less images (table \ref{nim}). To compare with the methods of \cite{Li2016} and \cite{Cho2015}, we first conduct our experiment on the same smaller sets of images that was used in \cite{Li2016}. Then, We test our method on the full set of ImageNet held-out categories that we downloaded from the ImageNet dataset.

Table \ref{selectedset} reports the co-localization of our method compares to \cite{Li2016} on the smaller ImageNet subset, and table \ref{fullset} reports the performance of our method on the ImageNet full subsets. The results show that our method significantly outperforms the competing methods in the smaller subset, with even higher accuracies for the full subset in Table \ref{fullset}. This result demonstrates that our method is robust to detect and localize truly unseen categories using previously learned CNN features.

\subsection{Qualitative results}

\begin{figure*}[t!]
\centering
\includegraphics[width=1.0\linewidth]{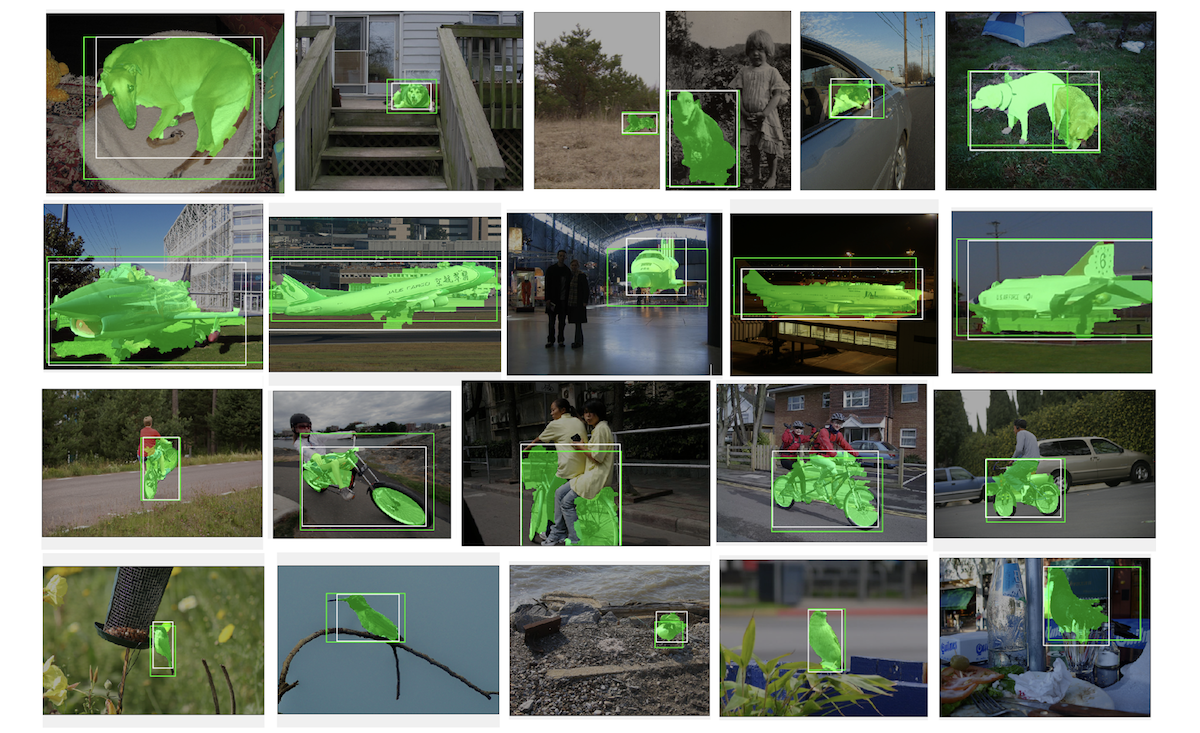}
   \caption{\textbf{Some example results of object co-localization with our CCFs and geodesic distance co-propagation.} The results show the bounding boxes that we generate match the ground truth bounding boxes very well, even when objects are not located centrally in the image. In addition, our co-localized object pixel-level regions (pixels colored in green) show well delineated shape in most cases. }
\label{fig:short}
\end{figure*}

\begin{figure*}[]
\begin{center}
\includegraphics[width=1.0\linewidth]{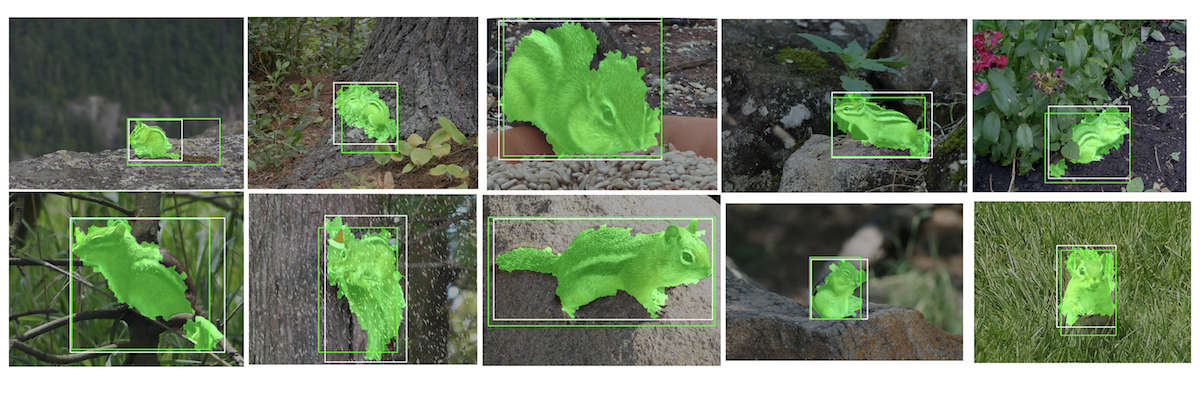}
\end{center}
\vspace{-0.7cm}
   \caption{\textbf{Some example results of our object co-localization on class "chipmunk"} from the six held-out ImageNet subset categories. The results show co-localized object pixel-level regions (pixels colored in green), the bounding boxes from our method (white), and ground truth bounding boxes (green). }
\label{fig:1}
\end{figure*}

We show some examples of our co-localization results in figure \ref{fig:short} and \ref{fig:1}. The results show that the bounding boxes generated by our proposed framework accurately match the ground truth bounding boxes. It is apparent that our results generate well-covered object regions, that has the potential to well delineate the objects in majority of the cases. The figures also show that the objects were able to be accurately co-localized with various sizes and locations.

\begin{figure}[h!]
\centering
\includegraphics[width=1.0\linewidth]{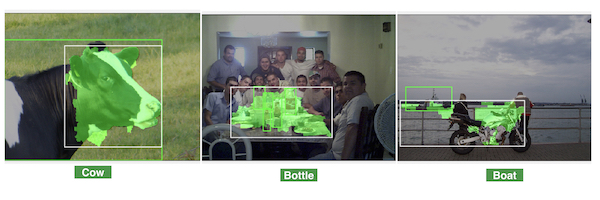}
   \caption{\textbf{Some failed examples of our approach.} While these examples did not have significant coverage with the ground truth boudning box in terms of CorLoc, the mistakes were still accurate, in the sense that the areas were detected but not the spatial extent. }
\label{fig:fail}
\end{figure}
%Notice that there are some images of ImageNet database that are unable to retrieve even though their bounding box annotations are available.

Figure \ref{fig:fail} illustrates three failure scenarios of our approach. While these three examples did not cover the ground truth bounding box sufficiently, but they were not far off. Some analysis suggests that these failures were due to some CCFs that are shared by multiple categories, and that the object boundaries may not have been strong enough (i.e. bottle and boat).

%\textbf{Implementation details} We use the outputs from the last convolutional layer of VGG-19 \cite{Simonyan14c} for this paper. We empirically use $K=5$-means for all the numbers reported. The propagation parameter $\mu$ is fixed at $0.5$. We generate the over-segmentation using geodesic k-means algorithm algorithm \cite{superpixel} and compute the soft boundary map by Structured Forests \cite{structured_forests}. 

\section{Conclusion}

In this work, we proposed a fully unsupervised 2-step approach for the problem of co-localization, that uses only positive images, and without any region or object proposals. Our method is motivated by human vision, people implicitly detect the common features of category examples to learn the representation of the class. We show that the identified category-consistent features can also act as an effective first-pass object detector. This idea is implemented by finding the group of CNN features that are highly and consistently activated by a given positive set of images. The result of this first step generates a rough but reliable object location, and acts as a single-shot object detector. Then, we aggregate the activations of the identified CCFs, and co-propagate their activations so that the activations over the true object region are boosted, while the activations over the background region are smoothed out. This effective activation refinement step allowed us to obtain accurately co-localized objects in terms of the standard CorLoc score with bounding boxes. We achieved new state-of-the-art performance on the three commonly used benchmarks. In the future, we plan to extend our method to generate unsupervised object co-segmentations. 

\FloatBarrier

\vspace{3mm}
{\small
\bibliographystyle{plainnat}
\bibliography{egbib}
}

\end{document}